\pdfoutput=1
\documentclass[11pt]{article}

\usepackage[final]{acl}

\usepackage{times}
\usepackage{latexsym}
\usepackage{algorithmic}
\usepackage{tabularx}
\usepackage{array}
\usepackage{tcolorbox}

\usepackage{graphicx}    % 用于插入图片
\usepackage[table,xcdraw]{xcolor}
\usepackage[T1]{fontenc}
\usepackage[utf8]{inputenc}

\usepackage{microtype}

\usepackage{inconsolata}

\usepackage[ruled]{algorithm2e}
\usepackage{makecell}
\usepackage{xcolor}
\usepackage{colortbl}
\usepackage{arydshln}

\definecolor{dashgray}{gray}{0.25}

\usepackage{multirow}
\usepackage{latexsym}
\usepackage{amsmath,amssymb,amsthm}
\usepackage{url}
\usepackage{microtype}

\usepackage{caption}
\usepackage{subcaption}
\usepackage{array}
\usepackage{longtable}

\usepackage{booktabs}

\title{From Sparse Representations to Behavioral Insights for Multimodal Depression Assessment}

\author{Guimin Hu$^{1}$, Zihao Song$^1$, Jiachen Luo$^{2}$, Jiayuan Xie$^{3}$, Ruichu Cai$^{1}$ \\
  $^1$Guangdong University of Technology\\
  $^2$Queen Mary University of London\\
  $^3$The Hong Kong Polytechnic University\\
\texttt{rice.hu.x@gmail.com}}

\begin{document}

\maketitle
\begin{abstract}
Multimodal depression assessment offers a promising approach to analyzing behavioral patterns associated with depression. However, existing methods often rely on dense and opaque multimodal representations, making it difficult to interpret the behavioral patterns underlying their predictions. In this work, we introduce BehavDep, a sparse factor-based framework that decomposes multimodal behavioral representations into sparse latent factors and associates them with behaviorally meaningful concepts through a semantic bridge. To address the mismatch between user-level annotations and heterogeneous video-level behaviors, BehavDep further learns video-level depression tendency scores under weak supervision and aggregates information across multiple observations for user-level assessment. Extensive experiments demonstrate that BehavDep achieves the best overall assessment performance while revealing complementary modality contributions, heterogeneous behavioral patterns across observations, and prediction responses to concept-level editing. These results show that BehavDep provides a structured and interpretable approach to analyzing multimodal behavioral representations for depression assessment.
\end{abstract}

\section{Introduction}
Depression is a prevalent mental health condition characterized by persistent changes in mood, cognition, and behavior. Its assessment typically relies on clinical interviews and standardized questionnaires that evaluate symptoms reported by individuals or observed by clinicians. However, depressive symptoms can also manifest through observable behavioral patterns, such as reduced facial expressiveness, changes in speech dynamics, and altered verbal communication. These behavioral manifestations provide a potential basis for computational approaches to depression assessment from naturally occurring behavioral data.

% \begin{figure}[t!]
%     \centering
%     \includegraphics[width=0.9\linewidth]{figs/teaser.pdf}
%     \caption{User-level depression assessment through sparse behavioral factors across a user's longitudinal video history.}
%     \label{fig:teaser}
% \end{figure}

Existing multimodal depression assessment methods primarily focus on predicting depression-related outcomes from multimodal behavioral signals. While these approaches have demonstrated promising predictive performance, they provide limited insight into how the underlying behavioral patterns are organized within the learned representations. A key challenge lies in the dense nature of multimodal representations. Different modalities and behavioral patterns are often encoded together in a shared latent space, making it difficult to disentangle and examine individual behavioral patterns and their associations with depression assessment. This motivates us to look beyond predictive performance and investigate the internal structure of multimodal behavioral representations. 

Furthermore, depression-related behavioral patterns may vary across observations from the same individual. In real-world mental health monitoring, an individual's behavioral data are often collected longitudinally, yet supervision is typically provided only at the user level. This granularity mismatch makes it difficult to characterize how behavioral patterns vary across individual observations and how such variation relates to depression assessment.
To address these challenges, we propose \textbf{BehavDep}, a sparse factor-based framework that transforms multimodal behavioral representations into interpretable behavioral insights for depression assessment. BehavDep first decomposes dense multimodal behavioral representations into sparse latent factors without imposing predefined semantic meanings. These factors are subsequently associated with depression-related behavioral concepts, providing a semantic basis for characterizing the behavioral patterns captured by the learned representation. To account for the heterogeneity of behavioral patterns across observations, BehavDep further estimates video-level depression tendency scores under user-level weak supervision, enabling the analysis of behavioral variation across individual observations. Together, these components provide a structured approach to analyzing multimodal behavioral representations and the behavioral patterns associated with depression assessment

The main contributions of this work are summarized as follows:
\begin{itemize}

\item We introduce a sparse factor-based representation learning framework that decomposes multimodal behavioral representations into sparse latent factors without imposing predefined semantic meanings, enabling more explicit analysis of the underlying behavioral patterns.

\item We ground the learned factors in clinically meaningful, depression-related behavioral concepts to provide a semantic interpretation of the factorized representation, while estimating video-level depression tendency scores to characterize heterogeneous behavioral patterns across individual observations.

\item Extensive experiments demonstrate that BehavDep achieves state-of-the-art performance while enabling interpretable analysis of multimodal behavioral patterns and their variations across individual observations.

\end{itemize}

\section{Related Work}

\subsection{Multimodal Depression Understanding}

Depression assessment has been increasingly explored using multimodal behavioral signals from online platforms and video-based datasets, where facial, vocal, and linguistic cues provide complementary information for identifying depressive states \cite{DBLP:conf/smm4h/PirinaC18,DBLP:conf/emnlp/YatesCG17,DBLP:journals/tkde/TaoYLWH24,zhou2022tamfn}. Recent approaches leverage transformer-based fusion and cross-modal representation learning to capture complex behavioral patterns and achieve promising performance. However, these models typically rely on dense latent representations, making it difficult to disentangle and examine the behavioral patterns captured by different components of the representation. Furthermore, depression-related behavioral patterns may vary across observations from the same individual, posing additional challenges for analyzing heterogeneous patterns under user-level weak supervision.

\subsection{AI for Mental Health Assessment}
Artificial intelligence has been increasingly applied to mental health analysis, leveraging advances in representation learning and large-scale pretrained models \cite{holzinger2017we,sun2021psyqa,moor2023foundation}. Recent studies explore knowledge-enhanced and instruction-tuned language models for mental health reasoning. MentaLLaMA \cite{yang2024mentallama} introduced a large-scale mental health instruction dataset and adapted LLaMA2 \cite{touvron2023llama} for interpretable mental health tasks, while MentalLLM \cite{xu2024mental} investigated prompting and fine-tuning strategies for mental health prediction.
Despite these advances, existing AI-based mental health systems often rely on implicit multimodal representations, making it difficult to disentangle and interpret the behavioral patterns captured by the models. Recent approaches have explored the use of clinically meaningful concepts to improve the interpretability of learned representations, but the relationship between these concepts and the underlying multimodal behavioral representations remains relatively underexplored.

\section{Method}

\subsection{Problem Formalization}
Formally, let $u_i=\{v_1,\ldots,v_N\}$ denote the $i$-th user with $N$ videos, where the $j$-th video is represented as
$v_j=\{f_1^j,f_2^j,\ldots,f_M^j\}$, consisting of $M$ frames. The associated label $y_i$ represents the mental state of user $u_i$. Since raw video and audio data are not directly processed by our model, we use pre-extracted visual and acoustic feature embeddings obtained from off-the-shelf feature extractors as the model inputs throughout training and inference.

\subsection{BehavDep}
\begin{figure}[t!]
    \centering
    \includegraphics[width=\linewidth]{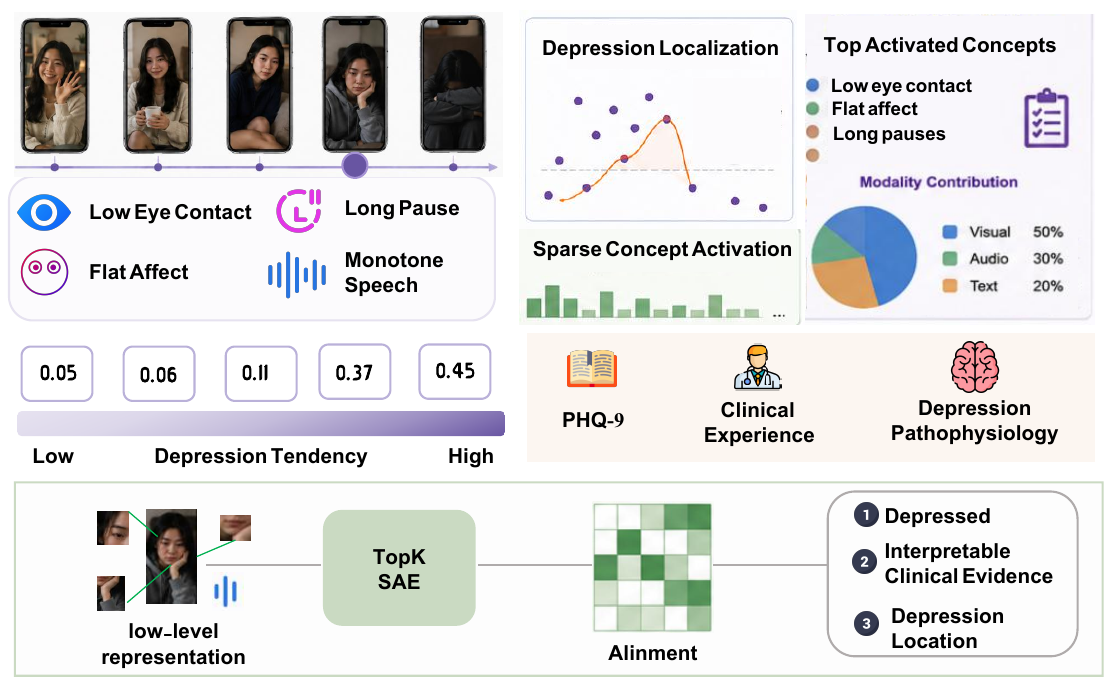}
    \caption{Overview of BehavDep.}
    \label{fig:overview}
\end{figure}

Figure~\ref{fig:overview} presents an overview of BehavDep, a clinically informed framework for interpretable multimodal depression assessment. Given an input video, BehavDep extracts multimodal behavioral representations and factorizes them into sparse latent factors without predefined semantics. The factors are then associated with a depression-related behavioral concept set curated from clinical literature and descriptions of depression-related behavioral manifestations, establishing a semantic bridge between latent factors and observable behaviors (e.g., low eye contact, long pauses, and slow speech). The complete concept set and supporting references are provided in Appendix \ref{app:depression_set}. Under user-level weak supervision, BehavDep further estimates video-level depression tendency scores, enabling fine-grained analysis of heterogeneous behavioral patterns and user-level depression assessment.
\subsubsection{Sparse Behavioral Representation}

\paragraph{Multimodal Behavioral Encoding.}
Given a user's video collection, each video is represented as a sequence of pre-extracted multimodal frame-level features. To capture temporal dependencies in multimodal behavioral signals, we employ a temporal encoder that projects frame-level features into a shared latent space and processes them with a Transformer encoder. Masked pooling is then applied to obtain a compact video-level representation. The resulting representations capture temporal behavioral patterns and serve as inputs for subsequent sparse factor learning. 

\paragraph{Sparse Latent Factor Discovery.}
Although multimodal behavioral encoders produce rich representations, these representations are typically dense and difficult to interpret at the component level. To obtain sparse latent factors, we pretrain a TopK Sparse Autoencoder (SAE) on the video-level representations. This factorization exposes structured latent patterns that facilitate subsequent analysis of behavioral characteristics relevant to depression assessment. Given the encoded video representation $\mathbf{x}$, the SAE encodes sparse latent factors and reconstructs the input as:
\begin{equation}
\mathbf{z}=f_{\mathrm{enc}}(\mathbf{x}), \quad \hat{\mathbf{x}}=f_{\mathrm{dec}}(\mathbf{z}),
\end{equation}
where TopK activation retains the top-ranked latent dimensions to encourage sparse representations. The SAE is optimized with reconstruction and sparsity objectives, and the pretrained encoder is subsequently used for behavioral concept association.

\paragraph{Behavioral Concept Association.}
Although the pretrained SAE discovers sparse latent factors, these factors remain semantically ambiguous, as individual dimensions do not have explicit semantic interpretations. To facilitate their interpretation, we construct a Depression-Related Behavioral Concept Set $\mathcal{C}=\{c_1,c_2,\ldots,c_K\}$, where each concept describes a behavioral pattern associated with depression, such as reduced facial expressiveness, monotonous speech, and social withdrawal.

Each behavioral concept is encoded into a semantic embedding using a pretrained text encoder:
\begin{equation}
\mathbf{t}_k=f_{\mathrm{text}}(c_k),
\end{equation}
where $\mathbf{t}_k$ denotes the semantic representation of the $k$-th behavioral concept. Given the sparse latent representation $\mathbf{z}$ obtained from the pretrained SAE, we project it into the same semantic space and measure its alignment with each concept embedding:
\begin{equation}
s_k=\cos(g(\mathbf{z}),\mathbf{t}_k),
\end{equation}
where $g(\cdot)$ denotes a learnable projection function. The resulting concept activation vector $\mathbf{s}=[s_1,\ldots,s_K]$ characterizes the association between the learned multimodal representation and the corresponding behavioral concepts for each video.

By associating the sparse latent factors of each video with behavioral concepts, BehavDep provides a semantic characterization of the learned factors and facilitates the analysis of behavioral patterns associated with depression.

\subsubsection{Concept-Aware Depression Assessment}

\paragraph{Concept Activation.}
After behavioral concept association, each video is represented by a concept activation vector $\mathbf{s}$, where each dimension corresponds to a depression-related behavioral concept. Unlike implicit latent representations, the concept activations provide a semantic space for characterizing the behavioral patterns captured by each observation. Specifically, a higher activation score indicates a stronger association between the observation and the corresponding behavioral concept. The resulting concept representation provides an interpretable intermediate space for characterizing multimodal behavioral patterns and supporting subsequent depression assessment.

Since depression-related behavioral patterns may vary across observations, directly averaging video-level representations may obscure differences in their contributions. To construct a user-level behavioral profile, we introduce a concept-aware aggregation strategy that adaptively weights videos according to their concept activation patterns.

Given the concept activation score of the $j$-th video $\mathbf{s}_j$, a learnable attention function estimates its relative contribution and then form concept aggregation across videos for user $u$:

\begin{equation}
\gamma_j=
\frac{\exp(h(\mathbf{s}_j))}
{\sum_{i=1}^{N_u}\exp(h(\mathbf{s}_i))},
\end{equation}

where $h(\cdot)$ denotes a learnable scoring function. $N_u$ denotes the video number of user $u$. The user-level concept profile is then obtained by weighted aggregation:

\begin{equation}
\mathbf{p}_u=\sum_{j=1}^{N_u}\gamma_j\mathbf{s}_j .
\end{equation}

The learned weights assign different contributions to individual observations based on their concept activation patterns, allowing the aggregation to emphasize more informative behavioral patterns while reducing the influence of less informative observations.

\paragraph{Concept-Guided Label Refinement.}

Although depression is assessed at the user level, depression-related behavioral patterns may vary across individual observations. Therefore, user-level annotations do not directly provide supervision for individual videos. To bridge this granularity gap, we derive video-level depression tendency scores from behavioral concept activations and learn them under user-level supervision.

For the $j$-th video, the concept activation vector $\mathbf{s}_j$ characterizes its associated behavioral patterns. A learnable mapping function generates a video-level depression tendency score, hereafter referred to as the depression soft label:
\begin{equation}
d_j=g(\mathbf{s}_j),
\end{equation}
where $d_j$ denotes the estimated depression tendency of the $j$-th video. The video-level tendency scores are then aggregated according to their contribution weights to obtain a user-level tendency:
\begin{equation}
d_u^{*}
=\sum_{j=1}^{N_u}\gamma_j d_j,
\end{equation}
where $\gamma_j$ denotes the contribution weight of the $j$-th video. The weights are estimated from the concept-aware representations and assign different contributions to individual observations, allowing more informative behavioral patterns to have greater influence on the user-level tendency.

The resulting user-level tendency is combined with the aggregated concept profile $\mathbf{p}_u$ for depression assessment:
\begin{equation}
\hat{y}_u=g([d_u^{*};\mathbf{p}_u]),
\qquad
\hat{y}_u\in\{0,1\},
\end{equation}

where $[;]$ denotes concatenation operation and $1$ denotes the depressive class. The user-level prediction is supervised by the ground-truth user label $y_u$, and the resulting loss provides weak supervision for learning the video-level tendency scores.

To obtain more stable video-level supervision, we further maintain a soft label bank that progressively refines the estimated tendency scores. Specifically, the newly estimated tendency scores are integrated with their historical values:
\begin{equation}
\mathbf{y}^{t+1}
=\alpha\mathbf{y}^{t}
+
\beta\mathbf{d}^{t},
\end{equation}
where $\mathbf{y}^{t}$ denotes the stored soft labels at iteration $t$, $\mathbf{d}^{t}=[d_1^{t},\ldots,d_{N_u}^{t}]$ denotes the newly estimated video-level tendency scores, and $\alpha,\beta$ control the contributions of historical and current estimates. The label bank is initialized with a temporal prior and iteratively updated using the learned tendency scores, providing more stable video-level supervision under user-level weak labels.

\paragraph{Longitudinal Behavioral Pattern Analysis.}

The video-level tendency score $d_i$ also provides a useful signal for examining behavioral patterns across a user's longitudinal video history. Videos with higher $d_i$ indicate stronger associations with the depressive class. By examining these videos together with their behavioral concept activations $\mathbf{s}_i$, we further analyze the behavioral patterns associated with higher depression tendency and their variation across observations.

\subsection{Experiments}

\subsubsection{Experimental Setting}
\paragraph{Datasets.}
\textbf{MUD3} \cite{DBLP:conf/mm/WangSZ025} is a multimodal user-level depression detection dataset with longitudinal social media videos, split into training, validation, and test sets with an 8:1:1 ratio. \textbf{DAIC-WOZ} \cite{gratch2014distress} is a clinical interview-based benchmark dataset providing synchronized audio, visual, and textual modalities with PHQ-8 scores. Detailed information is provided in the Appendix.

\begin{table}[t]
\centering
\small
\setlength{\tabcolsep}{3pt}
\renewcommand{\arraystretch}{1.1}
\begin{tabular}{lcccc}
\toprule
\textbf{Method} & \textbf{Acc.} & \textbf{F1} & \textbf{Pre.} & \textbf{Rec.} \\
\midrule
DepDetector$_{\mathrm{cat}}$ & 54.4 & 45.0 & 44.8 & 34.1 \\
DepDetector$_{\mathrm{han}}$ & 74.7 & 74.5 & 69.1 & 81.1 \\
\midrule
TAMFN$_{\mathrm{cat}}$ & 64.4 & 57.2 & 59.8 & 54.7 \\
TAMFN$_{\mathrm{han}}$ & 71.2 & 70.7 & 65.7 & 76.7 \\
\midrule
BiLSTM$_{\mathrm{cat}}$ & 73.6 & 71.1 & 71.5 & 72.2 \\
BiLSTM$_{\mathrm{han}}$ & 76.4 & 75.4 & 72.6 & 79.7 \\
\midrule
Transformer$_{\mathrm{cat}}$ & 73.7 & 70.3 & 73.3 & 68.7 \\
Transformer$_{\mathrm{han}}$ & 78.3 & 77.3 & 73.9 & 81.1 \\
\midrule
\textbf{BehavDep}
& $\mathbf{83.3}\scriptscriptstyle{\pm0.3}$
& $\mathbf{81.9}\scriptscriptstyle{\pm0.4}$
& $\mathbf{80.6}\scriptscriptstyle{\pm0.2}$
& $\mathbf{83.3}\scriptscriptstyle{\pm0.4}$ \\
\bottomrule
\end{tabular}
\caption{Performance comparison on the MUD3 dataset. Baseline results are adopted from \cite{DBLP:conf/mm/WangSZ025}. Results are reported as mean $\pm$ standard deviation over five independent runs.}
\label{tab:mud3_performance}
\end{table}

\begin{table}[t]
\centering
\small
\setlength{\tabcolsep}{0.8pt}
\renewcommand{\arraystretch}{1.0}
\begin{tabular}{lcccc}
\toprule
\multirow{2}{*}{\textbf{Method}}
& \multicolumn{2}{c}{\textbf{Normal}}
& \multicolumn{2}{c}{\textbf{Depressed}} \\
\cmidrule(lr){2-3}
\cmidrule(lr){4-5}
& \textbf{Pre.} & \textbf{F1}
& \textbf{Pre.} & \textbf{F1} \\
\midrule
ConvLSTM
& 83.1 & 35.1 & 30.1 & 45.2 \\
PerceiverIO
& 74.3 & 81.5 & 50.0 & 36.3 \\
AFABNet
& 77.4 & 83.6 & 62.2 & 45.8 \\
Qwen2-Audio-Instruct
& 79.1 & 79.1 & 50.0 & 51.3 \\
Mental-Perceiver
& 78.4 & 86.8 & 83.6 & 50.4 \\
\midrule
\textbf{BehavDep}
& $\mathbf{84.2}\scriptscriptstyle{\pm0.2}$
& $\mathbf{87.5}\scriptscriptstyle{\pm0.2}$
& $\mathbf{85.6}\scriptscriptstyle{\pm0.2}$
& $\mathbf{74.6}\scriptscriptstyle{\pm0.3}$ \\
\bottomrule
\end{tabular}
\caption{Performance comparison on the DAIC-WOZ dataset. Baseline results are reproduced from \cite{qin2025mental}.}
\label{tab:daic_woz}
\end{table}

\paragraph{Implementation Details.}

All experiments are implemented in PyTorch and conducted on a single NVIDIA GeForce RTX 4080 GPU. Training is performed in two stages. The sparse autoencoder is first trained for 20 epochs using a batch size of 256 and a learning rate of $1\times10^{-4}$, followed by 30 epochs of end-to-end optimization with a batch size of 64 and the same learning rate. Early stopping based on validation performance is employed to avoid overfitting. Please see Appendix \ref{app:more_implementation} for more implementation details.

\subsection{Main Results}
\paragraph{Result Analysis on MUD3}
As shown in Table~\ref{tab:mud3_performance}, BehavDep achieves the best performance across all evaluation metrics, improving accuracy and F1-score over the strongest baseline, Transformer$_{\text{han}}$, by 5.05 and 4.62 percentage points, respectively. We attribute these gains to the factorized and concept-guided representation learned by BehavDep. Rather than directly modeling dense multimodal representations through sequential modeling or feature-level fusion, BehavDep decomposes them into sparse latent factors and characterizes these factors through behavioral concepts, allowing the model to focus on more discriminative behavioral patterns. Moreover, the learned video-level depression tendencies provide finer-grained supervision for heterogeneous observations within the same user, complementing the user-level representation. The highest recall of 83.33\% further suggests that these complementary signals help the model identify depressive cases more effectively.

\paragraph{Result Analysis on DAIC-WOZ}
Table~\ref{tab:daic_woz} compares BehavDep with state-of-the-art methods on the DAIC-WOZ dataset. BehavDep achieves the best performance for both Normal and Depressed samples, with F1-scores of 0.87 and 0.74, respectively. Compared with the strongest baseline, Mental-Perceiver, BehavDep improves the F1-score by 1.2\% for Normal samples and 24.0\% for Depressed samples. These results demonstrate that the sparse factor-based representation can effectively capture discriminative behavioral patterns for depression assessment while providing a structured basis for their subsequent analysis.

\begin{figure}[t!]
    \centering
    \includegraphics[width=\linewidth]{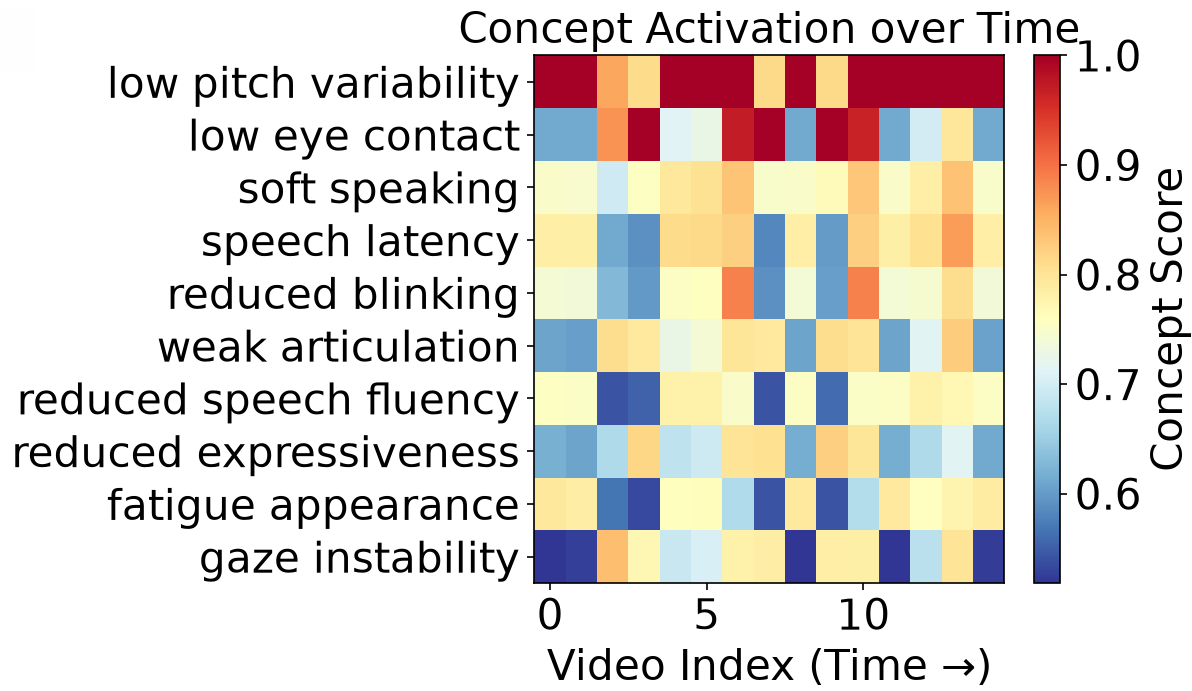}
    \caption{Temporal activation analysis of clinically grounded concepts for a depressed user. The heatmap illustrates the dynamic activation of depression-related behavioral concepts across video segments.}
    \label{fig:concept_analysis}
\end{figure}

\subsection{Behavioral Concept Analysis}

As shown in Figure~\ref{fig:concept_analysis}, BehavDep identifies several highly activated behavioral concepts, including low pitch variability, low eye contact, soft speaking, and reduced expressiveness. These concepts are consistent with behavioral manifestations commonly associated with depression, providing a semantic interpretation of the learned sparse factors. The observed factor--concept associations indicate that the sparse representation captures structured behavioral patterns rather than remaining as uninterpreted latent dimensions.

\begin{figure}[!t]
    \centering
    \begin{subfigure}{0.45\linewidth}
        \centering
        \includegraphics[width=\linewidth]{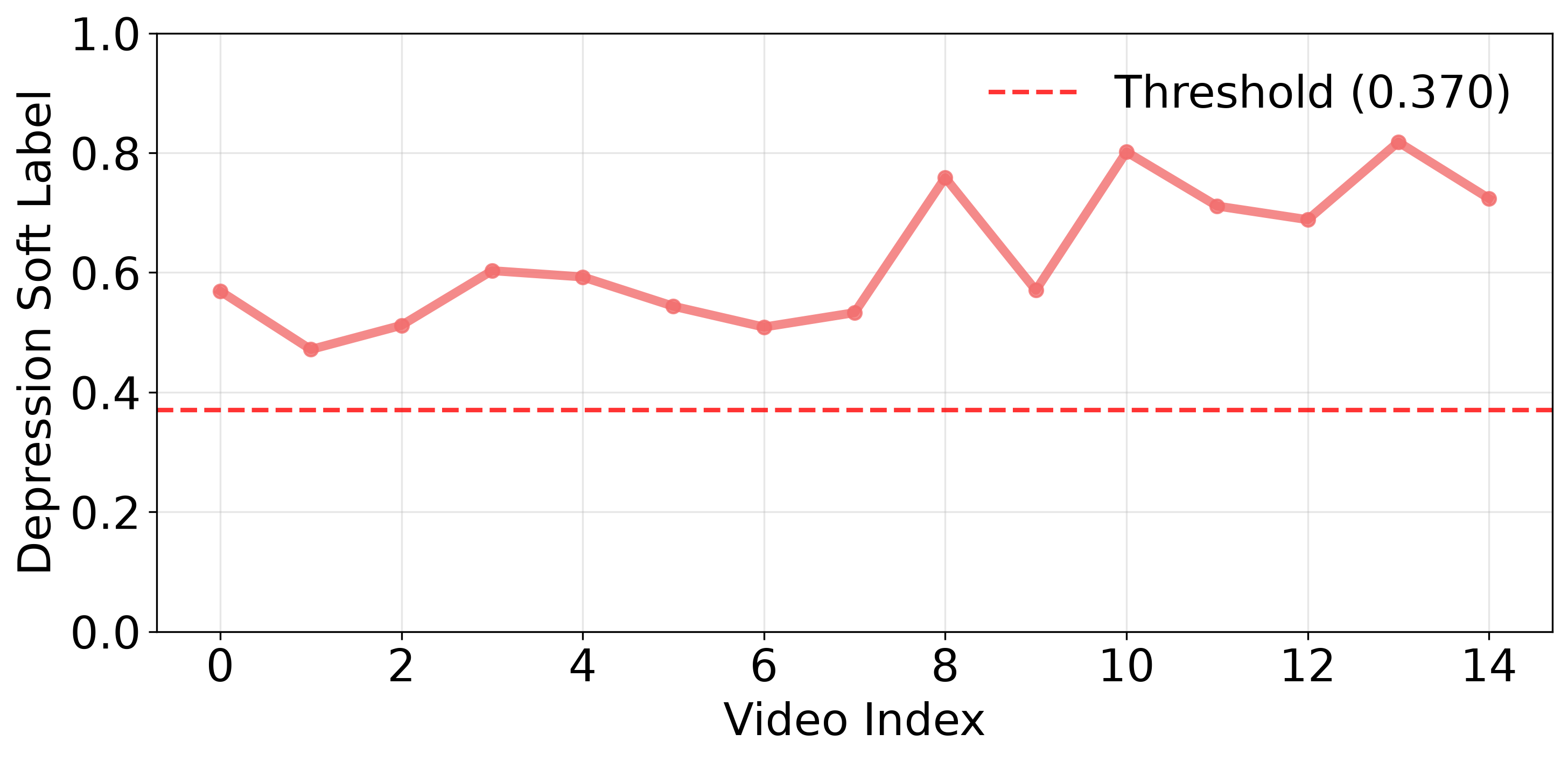}
        \caption{Depressed User}
        \label{fig:image1}
    \end{subfigure}
    \hfill
    \begin{subfigure}{0.45\linewidth}
        \centering
        \includegraphics[width=\linewidth]{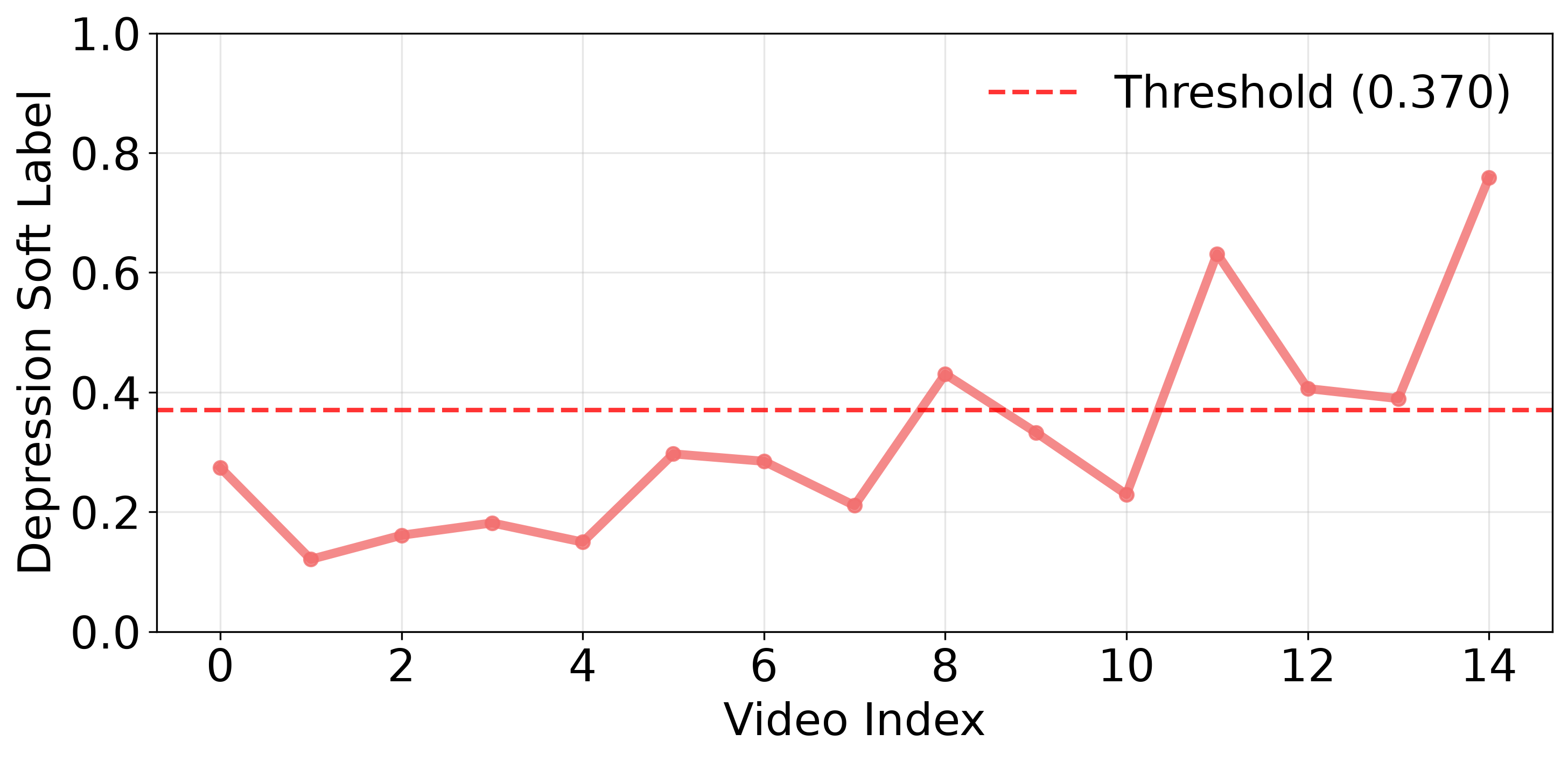}
        \caption{Depressed User}
        \label{fig:image2}
    \end{subfigure}
    
    \vspace{0.3cm}
    \begin{subfigure}{0.45\linewidth}
        \centering
        \includegraphics[width=\linewidth]{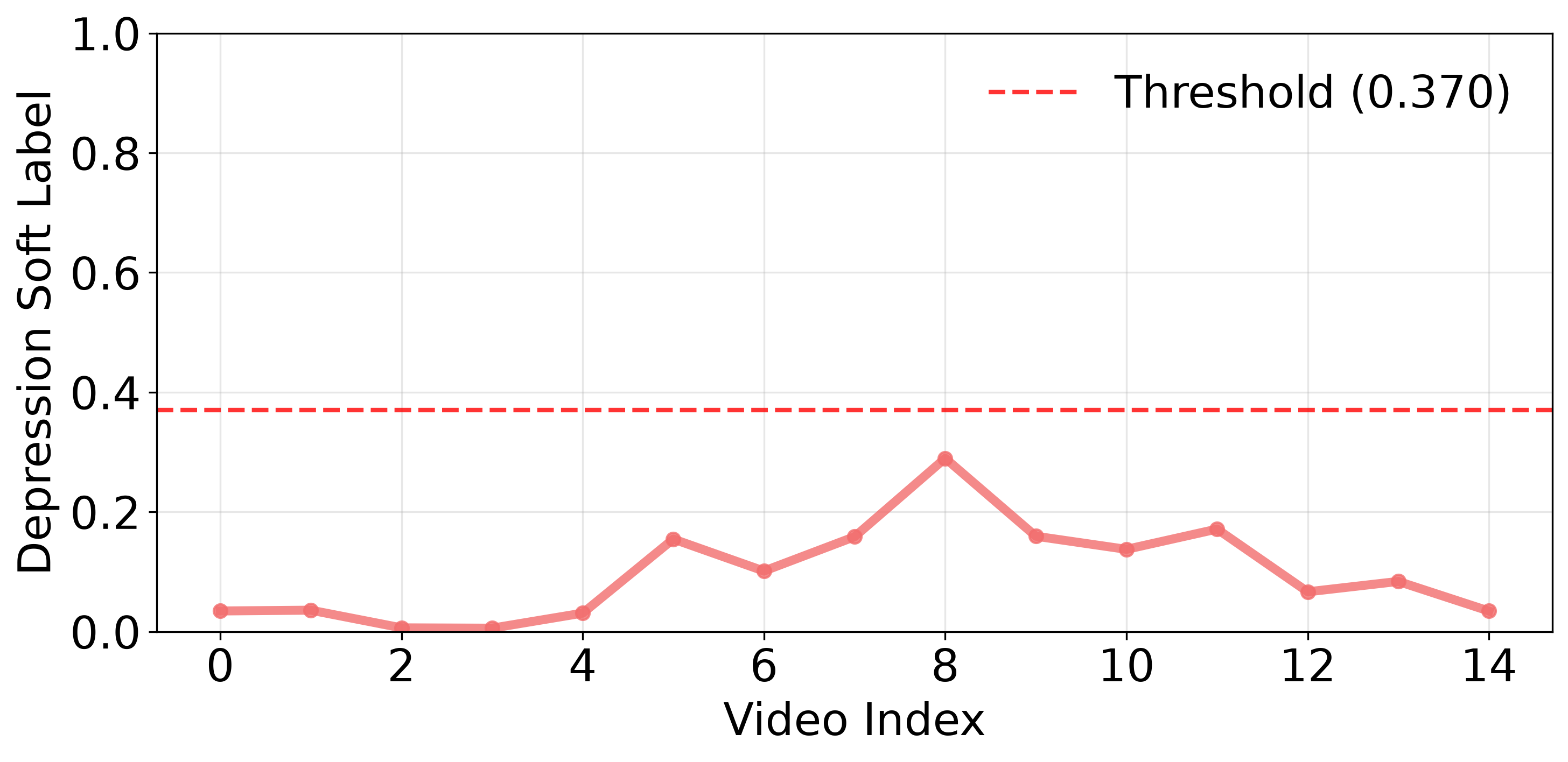}
        \caption{Non-Depressed User}
        \label{fig:image3}
    \end{subfigure}
    \hfill
    \begin{subfigure}{0.45\linewidth}
        \centering
        \includegraphics[width=\linewidth]{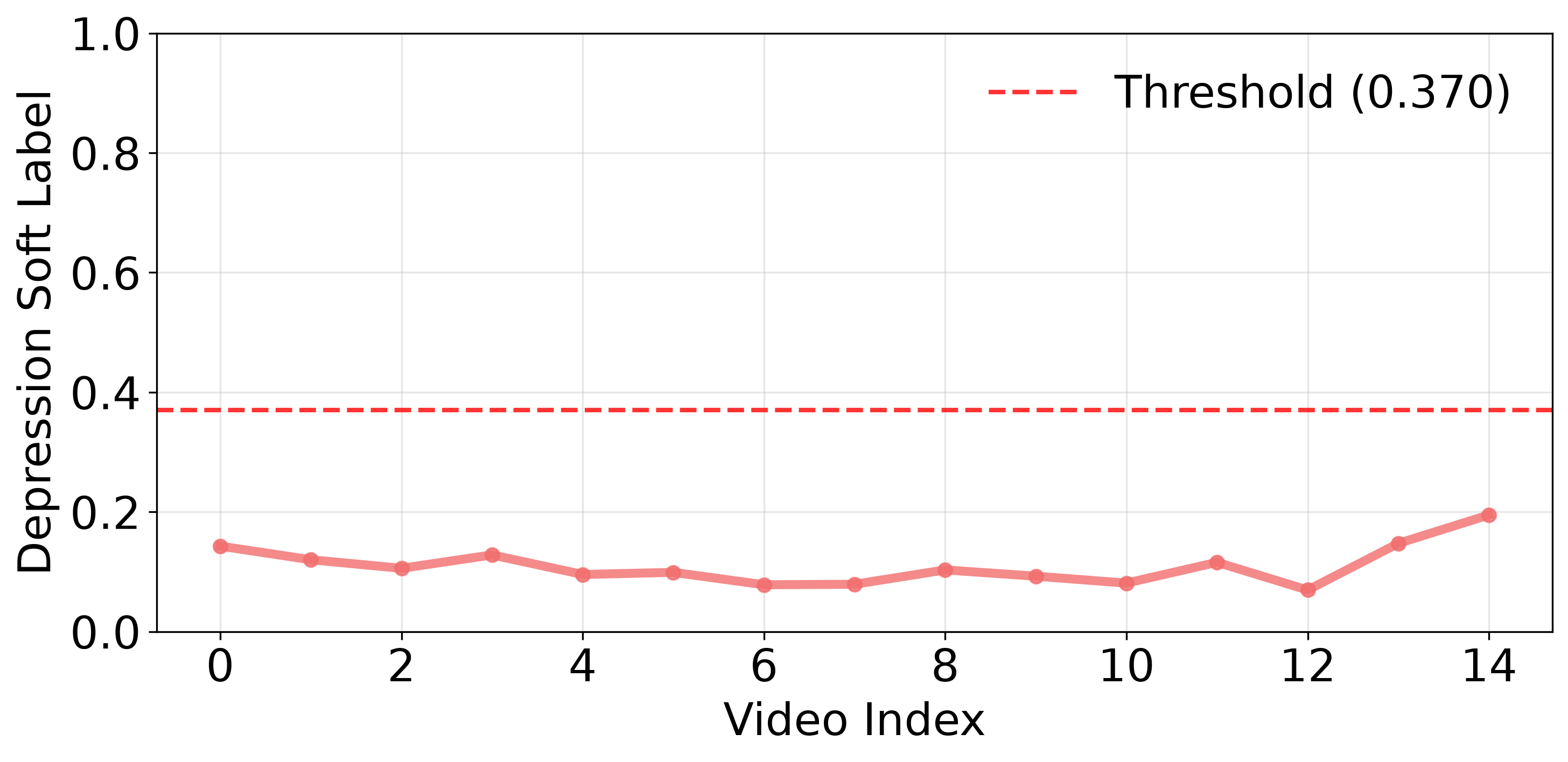}
        \caption{Non-Depressed User}
        \label{fig:image4}
    \end{subfigure}
    \caption{Video-level prediction trajectory. Dashed line denotes the threshold for depressive symptoms. }
    \label{fig:four_images}
\end{figure}

\begin{figure}[!ht]
\centering
\begin{tcolorbox}[
title=\textbf{Case: Concept-Guided Analysis},
colback=gray!3,
colframe=black!30,
boxrule=0.5pt,
arc=2mm,
width=0.95\linewidth
]

\textbf{Prediction:} Depressed

\textbf{Depression Probability:} 74.2\%

\textbf{Ground-Truth:} Depressed

\vspace{2mm}

\textbf{Behavioral Concept}

\textit{Acoustic Domain}

\begin{tabular}{ll}
Speech latency & 0.600\\
Long pauses & 0.555\\
Speech irregularity & 0.546\\
Reduced speech fluency & 0.541
\end{tabular}

\vspace{1mm}

\textit{Visual Domain}

\begin{tabular}{ll}
Reduced blinking & 0.588
\end{tabular}

\vspace{2mm}

\textbf{Behavioral Insight}

The detected behavioral patterns suggest
\emph{reduced speech energy}, \emph{psychomotor slowing},
and \emph{reduced facial expressiveness}, which are
commonly associated with depressive symptoms.

\end{tcolorbox}

\caption{Representative clinical prediction reports generated by BehavDep.}
\label{fig:case_1}
\end{figure}

The temporal activation patterns further reveal heterogeneous contribution dynamics across video segments. Concepts such as low pitch variability and low eye contact maintain relatively high activation across observations, whereas speech latency, reduced blinking, and gaze instability exhibit greater temporal variation. This suggests that different behavioral manifestations may emerge with different temporal patterns, with some expressed more consistently and others appearing intermittently across observations. Moreover, the complementary activation of multiple concepts indicates that the model captures diverse acoustic and visual behavioral patterns rather than relying on a single behavioral cue. Overall, these findings show that the learned sparse factors can be associated with semantically meaningful behavioral concepts while preserving the heterogeneous and temporally varying nature of multimodal behaviors.

\begin{table}[!t]
\centering
\small
\setlength{\tabcolsep}{3.5pt}
\renewcommand{\arraystretch}{1.1}
\begin{tabular}{lcccc}
\toprule
\textbf{Method} & 
\textbf{Acc.} & 
\textbf{F1} & 
\textbf{Pre.} & 
\textbf{Rec.} \\
\midrule
\textbf{BehavDep} 
& \textbf{83.3} & \textbf{81.9} & \textbf{80.6} & \textbf{83.3} \\

\midrule
w/o SAE     
& 77.2 & 76.5 & 71.4 & 78.1 \\
w/o Audio   
& 80.3 & 79.4 & 73.1 & 80.7 \\
w/o Video   
& 78.6 & 78.1 & 72.8 & 78.5 \\
w/o Concept 
& 77.3 & 76.4 & 71.9 & 78.5 \\
w/o Audio Concept 
& 78.7 & 77.6 & 72.1 & 78.9 \\
w/o Video Concept 
& 78.3 & 76.6 & 71.6 & 78.4 \\
\bottomrule
\end{tabular}
\caption{Ablation study on different components of BehavDep.}
\label{tab:ablation}
\end{table}

\begin{table*}[!t]
\centering
\small
\renewcommand{\arraystretch}{1.1}
\setlength{\tabcolsep}{4pt}

\begin{tabular}{lcccc}
\toprule
\textbf{Edited Concepts} &
\textbf{Modality} &
\textbf{Total Impact} &
\textbf{$\Delta$F1 (\%)} &
\textbf{Label Flip} \\
\midrule

\makecell[l]{Slow speech ($c_1$)}
& A
& 0.65
& -0.11
& Depressive $\rightarrow$ Depressive \\

\arrayrulecolor{dashgray}
\hdashline[0.5pt/2pt]
\arrayrulecolor{black}

\makecell[l]{Slow speech ($c_1$)\\
Long pauses ($c_2$)}
& A
& 1.14
& -0.25
& Depressive $\rightarrow$ Depressive \\

\arrayrulecolor{dashgray}
\hdashline[0.5pt/2pt]
\arrayrulecolor{black}

\makecell[l]{Flat facial affect ($c_6$)}
& V
& 0.72
& -0.27
& Depressive $\rightarrow$ Depressive \\

\arrayrulecolor{dashgray}
\hdashline[0.5pt/2pt]
\arrayrulecolor{black}

\makecell[l]{Flat facial affect ($c_6$)\\
Reduced facial movement ($c_7$)}
& V
& 1.36
& -1.84
& Depressive $\rightarrow$ Normal \\

\arrayrulecolor{dashgray}
\hdashline[0.5pt/2pt]
\arrayrulecolor{black}

\makecell[l]{Slow speech ($c_1$), Long pauses ($c_2$)\\
Flat facial affect ($c_6$), Reduced facial movement ($c_7$)\\
Low eye contact ($c_8$)}
& A,V
& 3.53
& -3.35
& Depressive $\rightarrow$ Normal \\

\arrayrulecolor{dashgray}
\hdashline[0.5pt/2pt]
\arrayrulecolor{black}

\makecell[l]{Long pauses ($c_2$), Flat facial affect ($c_6$)\\
Reduced facial movement ($c_7$), Low eye contact ($c_8$)\\
Psychomotor retardation ($c_{11}$), Emotional flattening ($c_{12}$)}
& A,V,S
& 3.81
& -4.58
& Depressive $\rightarrow$ Normal \\

\bottomrule
\end{tabular}
\caption{Counterfactual editing effects of clinically meaningful behavioral concepts. A, V, and S denote the audio, visual, and cross-modal concept categories, respectively.}
\label{tab:counterfactual}
\end{table*}

\subsection{Behavioral Pattern Analysis}

Beyond user-level depression assessment, the video-level tendency score $d_i$ provides an additional signal for examining how predictions vary across a user's longitudinal observations. Figure~\ref{fig:four_images} presents representative trajectories for two users with depression and two non-depressed users. The calculation of threshold for depressive symptoms is provided in Appendix \ref{app:more_implementation}. The variation in scores across observations indicates that the model does not assign a uniform depression tendency within a user, but instead captures heterogeneous prediction patterns across their video history.

To further characterize these observation-level differences, Figure~\ref{fig:case_1} presents a representative prediction case together with its associated behavioral concepts with the score. Both acoustic and visual concepts are activated, including speech latency, long pauses, and reduced blinking, illustrating how the factorized representation can be semantically characterized in terms of multimodal behavioral patterns. Overall, these results highlight two complementary capabilities of BehavDep: the video-level tendency score captures variation across observations, while factor-to-concept associations characterize the behavioral patterns underlying this variation. As the video-level scores are indirectly learned from user-level supervision, we regard this analysis as an exploratory characterization of behavioral heterogeneity rather than a definitive temporal characterization of depressive states.

\subsection{Ablation Study}

Table~\ref{tab:ablation} reports the impact of different components in BehavDep. Removing the SAE substantially reduces F1-score from 81.97\% to 76.52\%, demonstrating the importance of sparse factorization for capturing structured behavioral patterns. Removing concept guidance yields a similar drop to 76.35\%, highlighting the contribution of semantic guidance for interpreting depression-related behaviors. Modality ablations further demonstrate the complementary roles of acoustic and visual cues, as removing either modality consistently degrades performance. Removing modality-specific concepts also causes additional performance drops, suggesting that concept guidance captures complementary behavioral indicators across modalities. These results support the contributions of sparse factorization, concept guidance, and multimodal behavioral cues to BehavDep.

\begin{figure}[t!]
    \centering
    
    \begin{subfigure}[b]{0.48\linewidth}
        \centering
        \includegraphics[width=\linewidth]{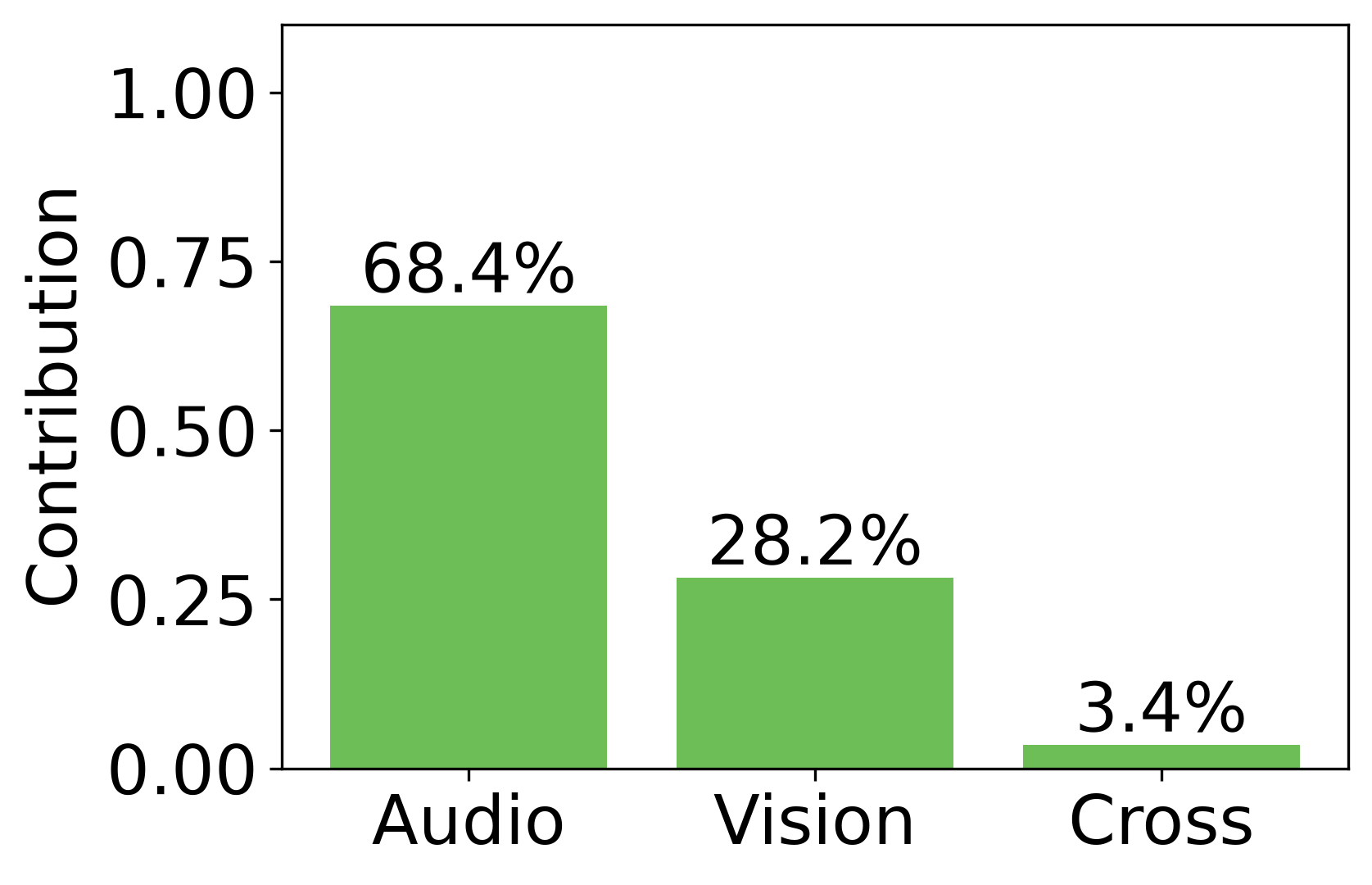}
        \caption{}
        \label{fig:modality_contribution_left}
    \end{subfigure}
    \hfill
    \begin{subfigure}[b]{0.46\linewidth}
        \centering
        \includegraphics[width=\linewidth]{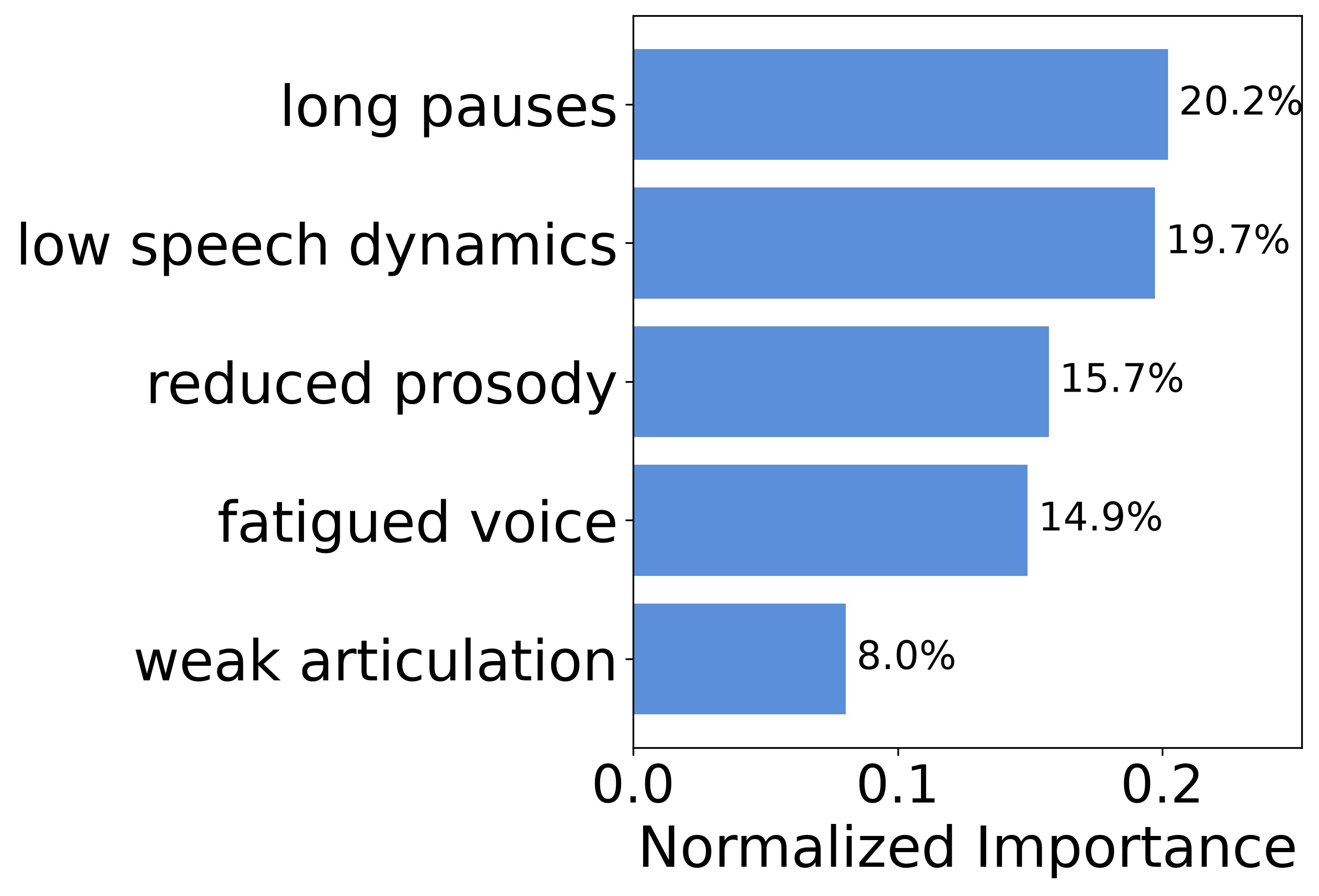}
        \caption{}
        \label{fig:modality_contribution_right}
    \end{subfigure}
    
    \caption{(a) Contribution analysis of different modalities and (b) acoustic feature importance}
    \label{fig:modal_analysis}
\end{figure}

\begin{figure*}[h!]
    \centering
    \includegraphics[width=\textwidth]{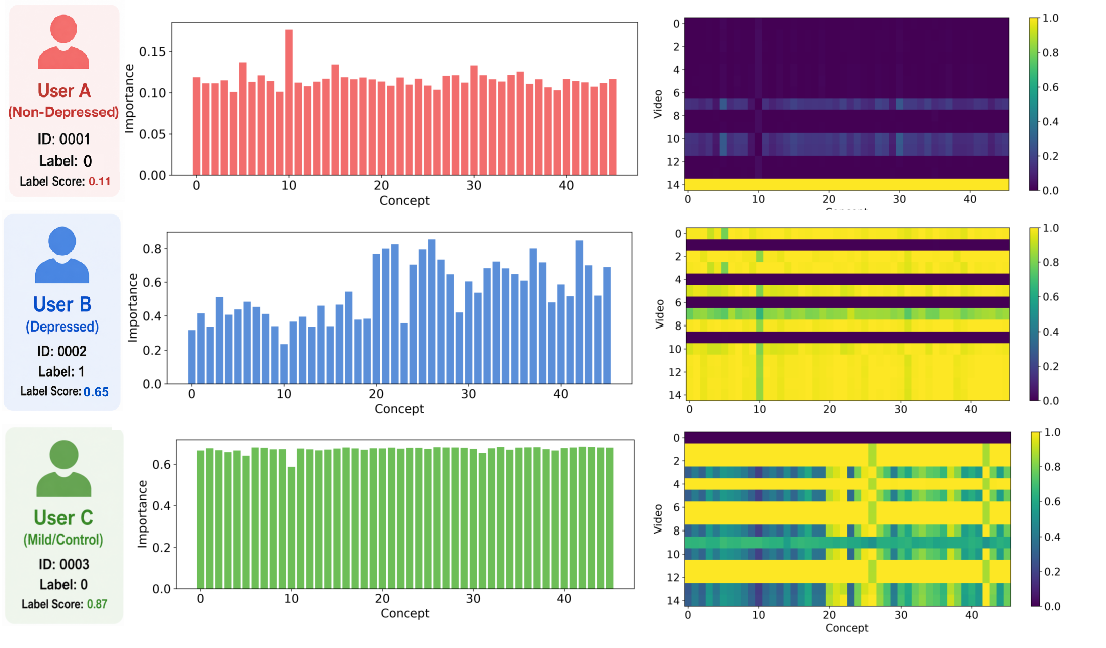}
    \caption{User-level case study involving three participants: User A is non-depressed, while Users B and C are diagnosed with depression.}
    \label{fig:user_analysis}
\end{figure*}

\section{Behavioral Concept Editing Analysis}

Counterfactual concept editing examines how changes in individual behavioral concepts affect model predictions. As shown in Table~\ref{tab:counterfactual}, editing a single concept generally produces only a modest reduction in the predicted depression tendency and rarely changes the final outcome. This suggests that no single behavioral cue dominates the assessment, and that the model can maintain its prediction when only one aspect of the behavioral profile is altered. In contrast, jointly editing multiple concepts across different behavioral dimensions produces substantially larger changes and consistently shifts the prediction from \textit{Depressive} to \textit{Normal}. The progressively larger prediction changes with additional concept edits indicate that the model responds to the accumulation of behavioral evidence rather than to an isolated cue. Notably, the edited concepts span different behavioral dimensions, such as acoustic and visual patterns, suggesting that their effects can accumulate across modalities and provide complementary evidence for the final prediction.

This contrast between isolated and joint editing provides insight into how the concept-aware representation supports depression assessment. Individual behavioral manifestations may provide only partial evidence, whereas coordinated changes across multiple concepts can substantially alter the overall behavioral profile and, consequently, the model's prediction. Thus, the counterfactual results suggest that BehavDep captures depression-related behavioral patterns as a combination of complementary concepts, rather than encoding them as a response to a single dominant behavioral indicator.

\subsection{Modal Contribution Analysis}
To investigate the underlying decision mechanism of the proposed model, we analyze both modality-level contribution and feature-level importance. As shown in Figure \ref{fig:modal_analysis}, the model primarily relies on acoustic information, with audio contributing the largest proportion among all modalities. This dominance is further supported by the feature importance analysis, where speech-related patterns such as long pauses and reduced speech dynamics are identified as the most influential cues. These results indicate that the model effectively captures behavioral characteristics reflected in temporal speaking patterns and vocal expressiveness, rather than relying on superficial correlations. Meanwhile, the visual modality provides complementary information, suggesting that multimodal integration helps enrich the representation of the target behavior. Overall, the consistency between modality-level contribution and feature-level importance demonstrates that the model learns meaningful and interpretable cues aligned with domain knowledge.

\subsection{User Study}

To examine the interpretability of the learned concept representations, we conduct a user-level case study involving three participants: User A is non-depressed, while Users B and C are diagnosed with depression. Figure~\ref{fig:user_analysis} visualizes concept importance (left) and concept activations across multiple videos (right). Concept importance jointly considers model attention and the semantic alignment between each concept embedding and the learned sparse representation (see Appendix for details).

User A shows consistently low concept activations across videos, with no prominent depression-related concepts. In contrast, Users B and C exhibit relatively consistent activations across multiple videos, while their concept profiles differ substantially despite sharing the same depression label. User B shows concentrated importance on several concepts, whereas User C exhibits a more distributed pattern. These observations illustrate both cross-video consistency and behavioral heterogeneity across users. Overall, the case study shows that concept representations expose fine-grained behavioral patterns underlying the model's depression assessment.

\section{Conclusion}

In this work, we presented BehavDep, a sparse factor-based framework for interpretable multimodal depression assessment. BehavDep decomposes dense multimodal behavioral representations into sparse latent factors and connects them with behaviorally meaningful concepts, enabling fine-grained characterization under coarse user-level supervision. Extensive experiments demonstrate improved assessment performance and reveal complementary modality contributions, heterogeneous and temporally varying behavioral patterns, and prediction responses to concept-level editing. Overall, BehavDep provides a structured approach to transforming opaque multimodal representations into interpretable behavioral evidence for depression assessment.

\subsection*{Limitations}
The longitudinal behavioral analysis should be interpreted as an exploratory analysis rather than a clinically validated characterization of depression-related behavior. The video-level tendency scores $d_i$ are indirectly learned from user-level supervision and therefore may be affected by the distribution of observations within each user's video history. Similarly, the associated behavioral concepts provide a semantic characterization of the learned representations but do not constitute clinically verified behavioral indicators. Future work could incorporate finer-grained annotations and expert evaluation to further validate the relationship between the identified behavioral patterns and depression-related assessments. BehavDep uses sparse factors and behaviorally meaningful concepts to provide interpretable descriptions of the behavioral patterns underlying its predictions. These model-generated characterizations are intended to provide a reference for fine-grained behavioral annotation, potentially alleviating the cost and difficulty of manually annotating behavioral patterns at the video level, rather than serving as clinical conclusions.

\section*{Ethical Considerations}

This study involves multimodal behavioral data for depression assessment, raising ethical considerations regarding privacy and potential misuse. All data are collected and processed in accordance with applicable ethical guidelines, with personally identifiable information removed or anonymized before analysis.

BehavDep is designed as an assistive research framework for analyzing multimodal behavioral patterns, rather than a clinical diagnostic system. The video-level depression tendency learned under weak supervision should not be interpreted as a clinical diagnosis or definitive measure of an individual's mental health status, and model outputs should not replace professional assessment.

We acknowledge potential risks related to algorithmic bias, model uncertainty, and misinterpretation.  Any practical deployment should incorporate appropriate human oversight and comply with relevant ethical standards, with particular attention to privacy, fairness, informed consent, and the consequences of erroneous predictions.

\bibliography{acl_latex}

\appendix

\section{Depression Concept Set}
\label{app:depression_set}
To establish a clinically grounded representation for depression assessment, the proposed Depression Concept Set is constructed based on clinically recognized depressive symptoms described in the DSM-5 ~\cite{edition2013diagnostic,buyukdura2011psychomotor}. We summarize representative behavioral concepts across modalities by reviewing studies on depressive symptoms, affective behaviors, and observable depression markers.
\begin{figure}[ht]
\centering

\fbox{
\begin{minipage}{0.9\linewidth}

\textbf{Concept Bank for Depression Understanding}

\vspace{0.2cm}

\textbf{Acoustic Concepts:} \\
slow speech,
long pauses,
reduced speech,
low vocal energy,
monotone speech,
low pitch variability,
hesitant speaking,
breathy voice,
weak articulation,
speech irregularity,
soft speaking,
reduced verbal output,
speech latency,
slowed response timing,
voice instability,
reduced speaking engagement,
flat vocal affect,
low speech dynamics,
fatigued voice

\vspace{0.2cm}

\textbf{Visual Concepts:} \\
flat facial affect,
reduced facial movement,
low eye contact,
reduced blinking,
downward gaze,
head lowering,
reduced movement,
facial tension,
emotion suppression,
reduced expressiveness,
gaze instability,
slumped posture,
psychomotor slowing,
low behavioral activation,
facial asymmetry,
micro-expression reduction,
reduced smiling,
avoidant gaze,
visual disengagement,
fatigue appearance

\vspace{0.2cm}

\textbf{Cross-modal Concepts:} \\
psychomotor retardation,
emotional flattening,
behavioral withdrawal,
social disengagement,
cognitive fatigue,
low interpersonal engagement

\end{minipage}
}
\caption{Structured concept bank.}
\label{fig:concept_bank_structured}
\end{figure}

The concept bank is organized into three categories: acoustic concepts, visual concepts, and cross-modal concepts. The acoustic concepts are derived from psychological evidence regarding speech-related changes in depression, capturing characteristics such as slowed speech, prolonged pauses, reduced vocal energy, monotonic prosody, decreased speech fluency, and diminished verbal engagement. These concepts reflect alterations in speech production, emotional expressiveness, and communicative motivation frequently reported in individuals with depressive symptoms. The visual concepts are constructed based on documented nonverbal behavioral indicators associated with depression, including reduced facial expressiveness, limited eye contact, downward gaze, decreased facial movement, slumped posture, and psychomotor slowing. These concepts represent changes in emotional display, social interaction, and behavioral activation that can be observed through facial and body cues. In addition, we introduce cross-modal concepts to capture high-level depressive behaviors that typically emerge from the integration of multiple behavioral channels, such as psychomotor retardation, emotional flattening, behavioral withdrawal, and social disengagement. These concepts provide a more comprehensive description of depression-related patterns beyond individual modality-specific observations. By grounding the concept bank in established psychological findings, we aim to bridge low-level multimodal signals and high-level clinical interpretations. This literature-guided design enables the model to perform depression understanding through semantically meaningful concepts, improving interpretability while maintaining alignment with psychological knowledge. Table \ref{tab:clinical_concept_sources} presents representative clinical concepts along with their corresponding supporting literature.

\begin{table*}[t]
\centering
\small
\setlength{\tabcolsep}{4pt}
\renewcommand{\arraystretch}{1.08}

\begin{tabular}{l p{0.27\textwidth} p{0.52\textwidth}}
\toprule
\textbf{Clinical Concept} &
\textbf{Clinical Basis} &
\textbf{Observable Manifestation} \\
\midrule

Low eye contact
& Reduced social engagement \cite{buyukdura2011psychomotor}
& Reduced gaze toward the interlocutor \\

Flat facial affect
& Diminished emotional expression \cite{buyukdura2011psychomotor}
& Limited facial and emotional expression \\

Long pauses
& Psychomotor slowing \cite{bennabi2013psychomotor}
& Prolonged pauses or delayed responses \\

Monotone speech
& Diminished emotional expression \cite{bennabi2013psychomotor}
& Reduced pitch and prosodic variation \\

Soft speaking
& Reduced energy \cite{almaghrabi2023bio}
& Low vocal intensity and speaking volume \\

Slow speech
& Psychomotor slowing \cite{buyukdura2011psychomotor}
& Reduced speech rate \\

Reduced movement
& Psychomotor retardation \cite{paquet2022psychomotor}
& Limited head, facial, or upper-body movement \\

Reduced speech
& Reduced verbal activity \cite{almaghrabi2023bio}
& Reduced amount of verbal output or shorter utterances \\

\bottomrule
\end{tabular}

\caption{Behavioral concepts, their literature bases, and corresponding observable manifestations used in BehavDep.}
\label{tab:clinical_concept_sources}
\end{table*}

% \begin{table*}[t] \centering \small \setlength{\tabcolsep}{4pt} \renewcommand{\arraystretch}{1.08} \begin{tabularx}{\textwidth}{l l X} \toprule \textbf{Clinical Concept} & \textbf{Clinical Basis} & \textbf{Observable Manifestation} \\ \midrule Low eye contact & Reduced social engagement \cite{buyukdura2011psychomotor} & Reduced gaze toward the interlocutor \\ Flat affect & Diminished emotional expression \cite{buyukdura2011psychomotor} & Limited facial and emotional expression \\ Long pauses & Psychomotor slowing \cite{bennabi2013psychomotor} & Prolonged pauses or delayed responses \\ Monotone speech & Diminished emotional expression \cite{bennabi2013psychomotor} & Reduced pitch and prosodic variation \\ Soft speaking & Reduced energy \cite{almaghrabi2023bio} & Low vocal intensity and speaking volume \\ Slow speech & Psychomotor slowing \cite{buyukdura2011psychomotor} & Reduced speech rate \\ Reduced movement & Psychomotor retardation \cite{paquet2022psychomotor} & Limited head, facial, or upper-body movement \\ Reduced speech & Reduced verbal activity \cite{almaghrabi2023bio} & Reduced amount of verbal output or shorter utterances \\ \bottomrule \end{tabularx} \caption{Clinical concepts, their clinical bases, and corresponding observable behavioral manifestations used for clinical grounding in BehavDep.} \label{tab:clinical_concept_sources} \end{table*}

\section{Baselines}
To comprehensively evaluate BehavDep, we compare it with representative baselines, covering traditional machine learning approaches, multimodal neural architectures, and recent state-of-the-art depression assessment models:
\begin{itemize}
    \item Dep~\cite{yoon2022d} a Transformer-based model that extracts unimodal temporal features and aligns modalities through cross-attention and concatenation.
    \item TAMFN~\cite{zhou2022tamfn} a multimodal model that performs early feature fusion by integrating intermediate unimodal representations through temporal attention-based cross-modal fusion.
    \item BiLSTM~\cite{huang2015bidirectional} a bidirectional LSTM-based model that jointly processes two modalities and aggregates temporal information through an attention mechanism.
    \item Trans~\cite{vaswani2017attention} a Transformer-based model that captures temporal dependencies by aggregating information across the temporal dimension.
    % \item SVM~\cite{pedregosa2011scikit} a robust shallow classifier that performs binary classification efficiently using the kernel trick to project data into a higher-dimensional feature space.
    % \item RandomForest~\cite{pedregosa2011scikit} an ensemble learning method that performs classification by aggregating the predictions of multiple decision trees.
    % \item XGBoost~\cite{chen2016xgboost} an efficient gradient boosting framework for classification based on optimized distributed tree boosting.
    % \item NUSD~\cite{wang2023non} a deep ECAPA-TDNN-based model enhanced with speaker disentanglement through non-uniform adversarial speaker identity loss maximization.
    \item ConvLSTM~\cite{wei2022multi} a convolutional bidirectional LSTM with a sub-attention mechanism for modeling heterogeneous information.
    \item PerceiverIO~\cite{jaegle2021perceiver} a general-purpose architecture for multimodal learning with a fully attention-based design and a flexible querying mechanism.
    \item AFABNet~\cite{xu2024attention} an attention-based acoustic feature fusion network that combines four complementary acoustic features for depression detection.
    \item Qwen2-Audio-Instruct~\cite{chu2024qwen2} a 7B audio-language model based on Qwen that supports diverse audio inputs and performs speech understanding and instruction-following through natural language responses.
    \item Mental-Perceiver~\cite{qin2025mental} a multimodal deep learning model that leverages audio and textual information for anxiety and depression detection.
\end{itemize}

\begin{figure}[ht!]
    \centering
    \includegraphics[width=0.6\linewidth]{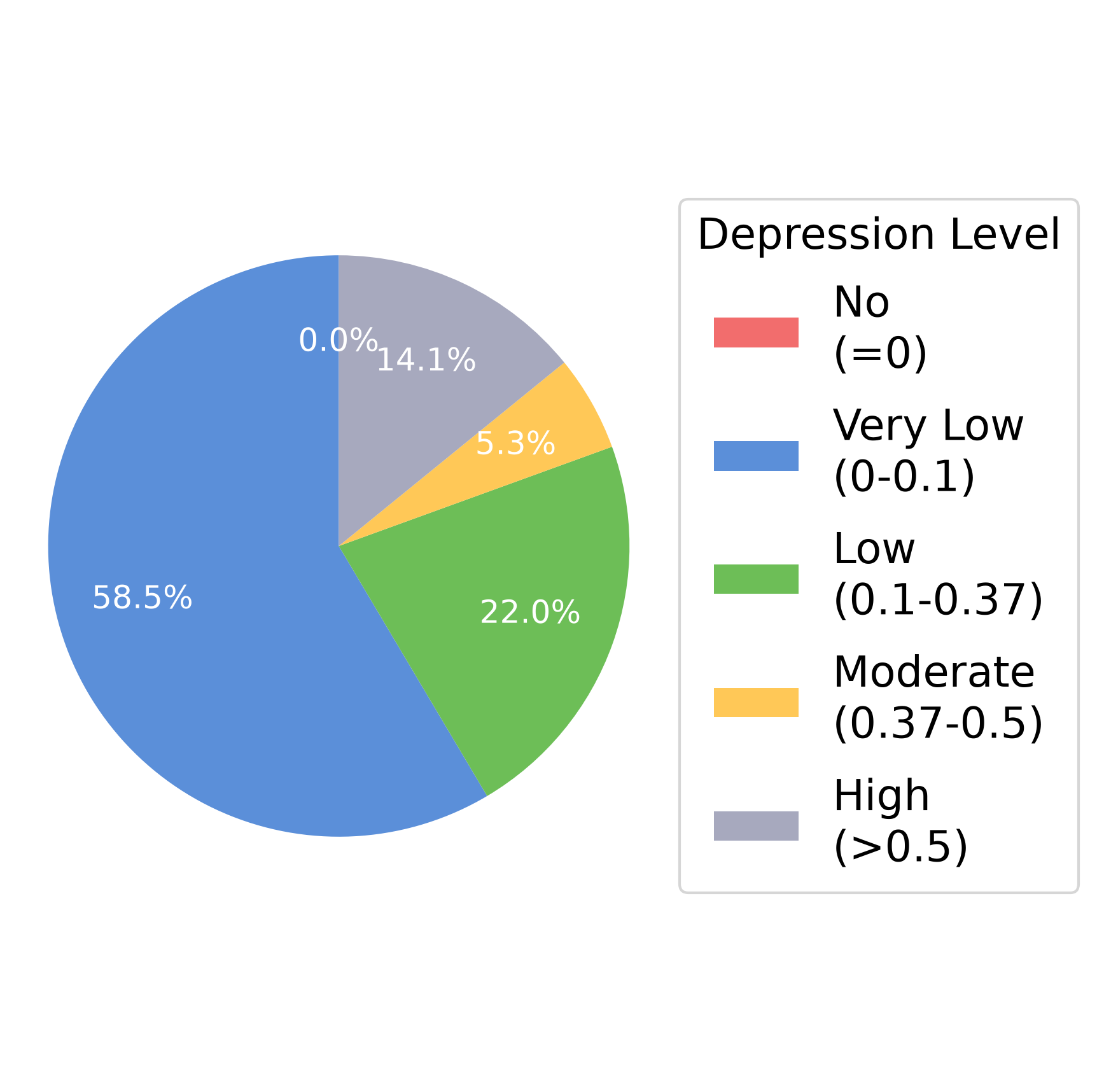}
    \caption{Distribution of Depression Evidence.}
    \label{fig:depression_distribution}
\end{figure}

\section{Observation-level Behavioral Heterogeneity}

Figure~\ref{fig:depression_distribution} shows the distribution of the learned video-level depression tendency scores across observations from users with depression. The scores exhibit a highly uneven distribution: 58.5\% of the 3,192 video segments receive scores below 0.1, while only 14.1\% receive scores above 0.5. This indicates that depression-related tendencies are not uniformly distributed across a user's observations, with stronger tendencies concentrated in a relatively small subset of segments.

This observation supports the need for video-level tendency estimation under user-level supervision. Rather than assuming that all observations from a depressed user contribute equally to the assessment, BehavDep allows individual observations to exhibit different levels of depression tendency. Combined with the associated behavioral concepts, these scores further provide a basis for characterizing the behavioral patterns represented in observations with different depression tendencies.

\section{Intra-User Variability of Depression Tendency}
To characterize the variability of depression-related signals across observations, we visualize the video-level depression soft-label trajectories of representative non-depressed and depressed users. Each point denotes the predicted depression tendency of an individual video. As shown in Figure~\ref{fig}, the predicted scores vary substantially across videos within the same user, rather than remaining at a consistent level. This variation indicates that depression-related behavioral signals can differ considerably across observations, suggesting that a single video may not fully characterize a user's overall mental state.

These observations motivate the use of user-level aggregation to integrate multiple heterogeneous video instances rather than relying on individual video predictions. They also highlight the importance of modeling intra-user variability when learning video-level depression tendencies, as different observations from the same user may exhibit substantially different levels of depression-related behavioral evidence.

\begin{figure}[!ht]
    \centering
    \begin{subfigure}{0.45\linewidth}
        \centering
        \includegraphics[width=\linewidth]{figs/27_user_413.png}
        \caption{Non-Depressed User}
        \label{fig:image3}
    \end{subfigure}
    \hfill
    \begin{subfigure}{0.45\linewidth}
        \centering
        \includegraphics[width=\linewidth]{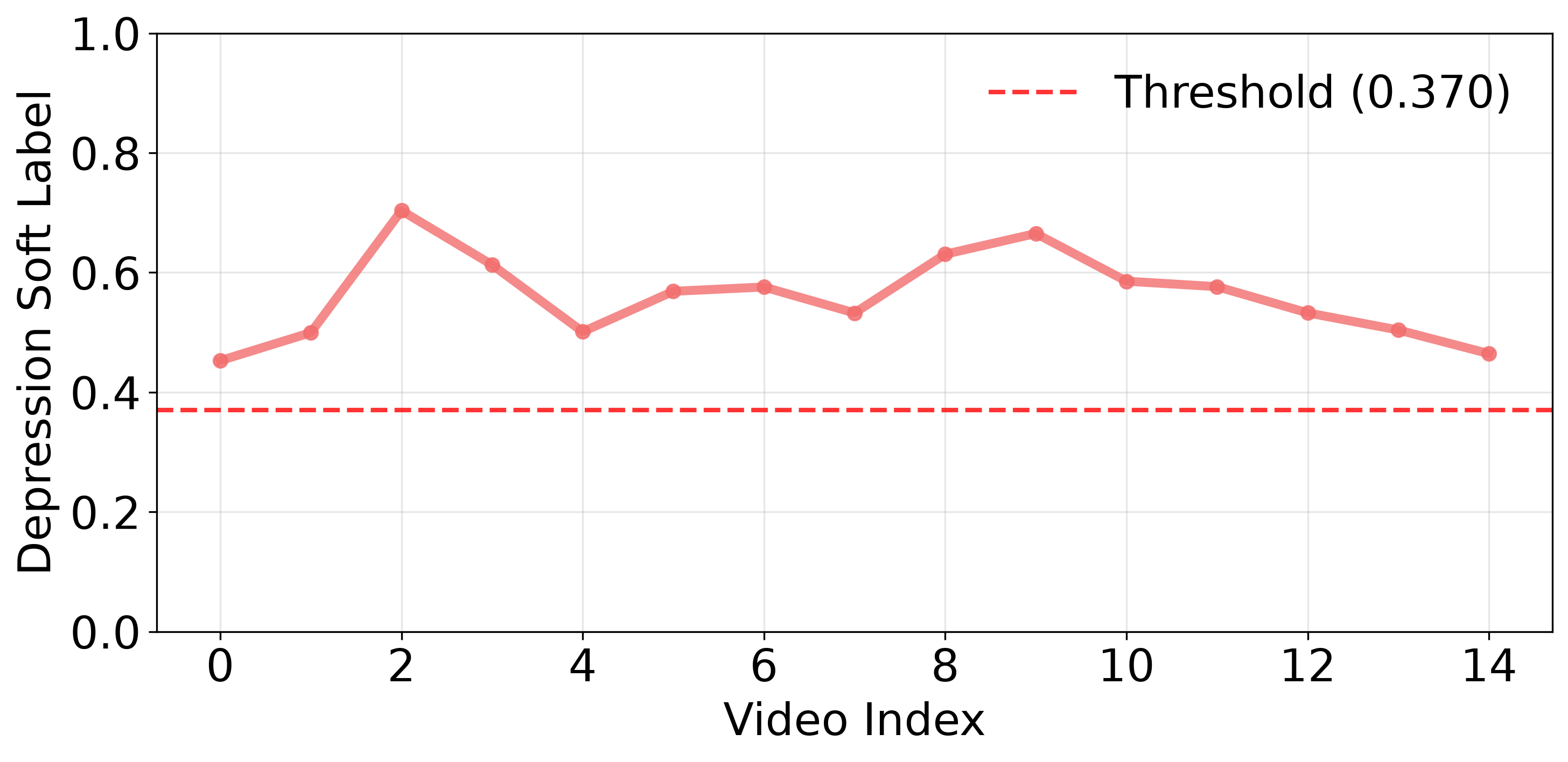}
        \caption{Depressed User}
        \label{fig:image4}
    \end{subfigure}

            \vspace{0.3cm}
    \begin{subfigure}{0.45\linewidth}
        \centering
        \includegraphics[width=\linewidth]{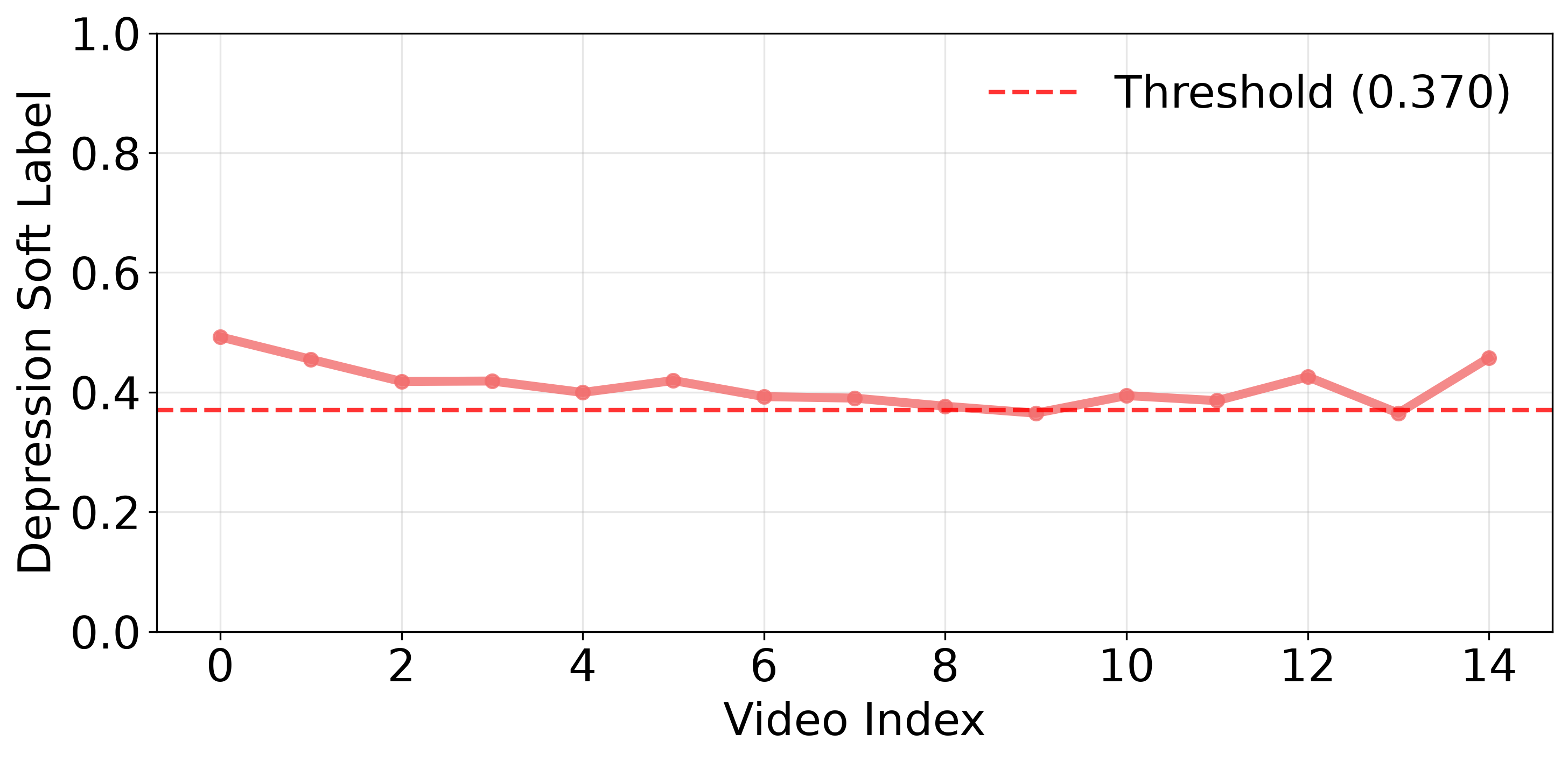}
        \caption{Depressed User}
        \label{fig:image3}
    \end{subfigure}
    \hfill
    \begin{subfigure}{0.45\linewidth}
        \centering
        \includegraphics[width=\linewidth]{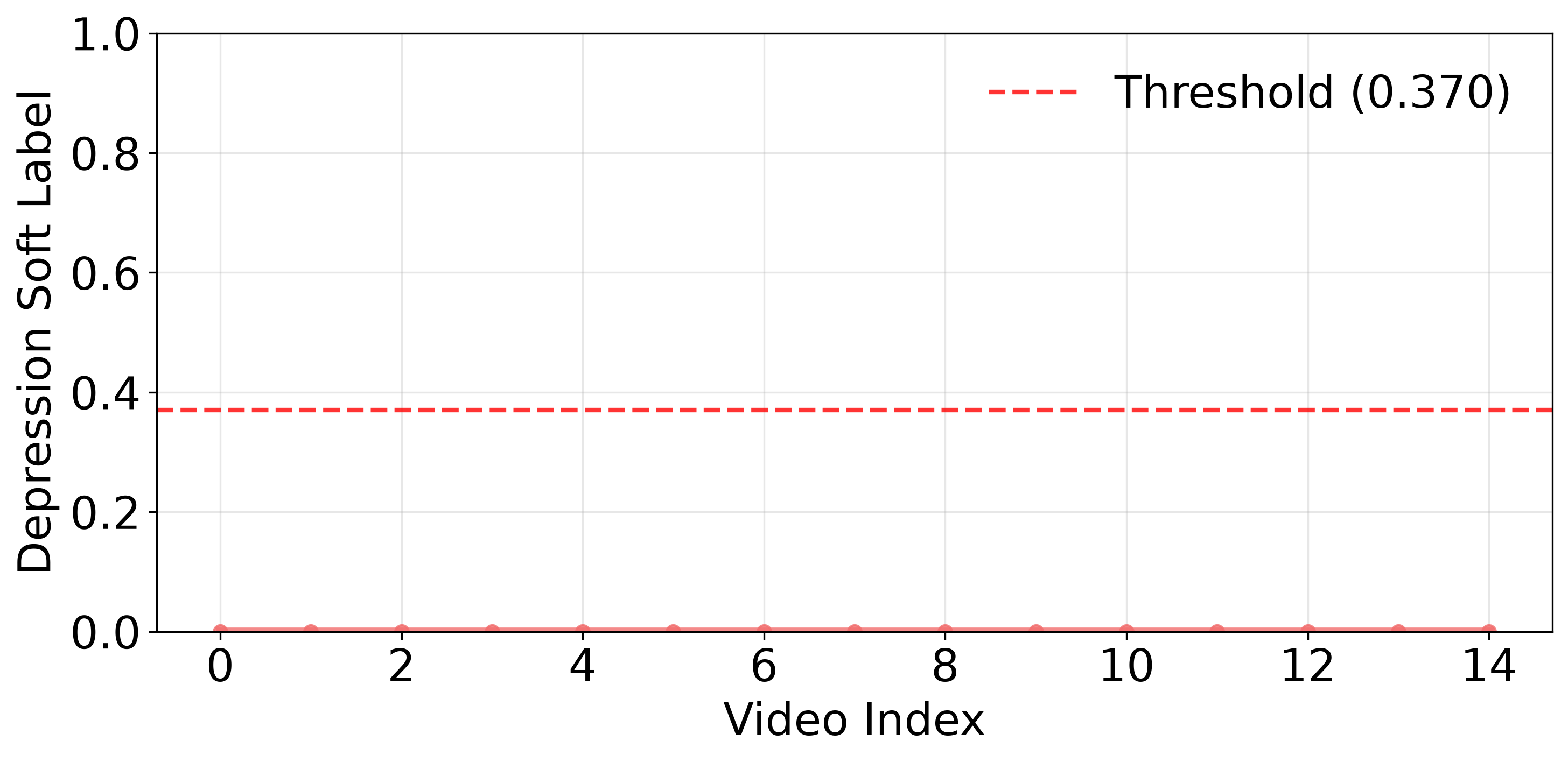}
        \caption{Non-Depressed User}
        \label{fig:image4}
    \end{subfigure}

    \caption{Temporal prediction trajectory of BehavDep.}
    \label{fig:four_images_appendx}

\end{figure}

\section{Computation of Concept Importance}

To quantify the contribution of each clinically grounded concept, we define concept importance by integrating two complementary signals: (1) the model attention indicating how much the prediction relies on each concept, and (2) the semantic alignment score measuring how well the learned sparse representation encodes the corresponding clinical concept.

Given the sparse representation $\mathbf{z}$ and concept embedding $\mathbf{c}_k$, the semantic alignment score for the $k$-th concept is computed as:

\begin{equation}
s_k =
\frac{\mathbf{z}^{T}\mathbf{c}_k}
{\|\mathbf{z}\|_2\|\mathbf{c}_k\|_2},
\end{equation}

where $s_k$ measures the correspondence between the learned sparse feature and the clinical concept. Meanwhile, the attention weight $\alpha_k$ represents the model's reliance on the $k$-th concept during depression assessment. The final concept importance is obtained by combining these two factors:

\begin{equation}
I_k = \alpha_k \cdot s_k,
\end{equation}

where $I_k$ denotes the importance of the $k$-th concept, reflecting its overall contribution to the depression prediction.

\begin{figure}[!ht]
\centering

% Case 1
\begin{tcolorbox}[
title=\textbf{Case 1: Acoustic-based Prediction Report},
colback=gray!3,
colframe=black!30,
boxrule=0.5pt,
arc=2mm,
width=0.95\linewidth
]

\textbf{Prediction:} Depressed

\textbf{Depression Probability:} 60.5\%

\vspace{2mm}

\textbf{Behavioral Evidence}

\begin{tabular}{ll}
Speech latency & 0.571\\
Voice instability & 0.509\\
Low speech dynamics & 0.508\\
Long pauses & 0.505\\
Reduced speech fluency & 0.494
\end{tabular}

\vspace{2mm}

\textbf{Behavioral Insight}

The detected behavioral pattern suggests
\emph{reduced speech energy}, \emph{long pauses},
and \emph{psychomotor slowing}, which are commonly
associated with depressive symptoms.

% \vspace{2mm}

% \textbf{Recommendation}

% Further clinical evaluation is recommended.

\end{tcolorbox}

\caption{Representative behavioral insight report generated by BehavDep, illustrating how sparse multimodal behavioral factors are translated into interpretable evidence for depression assessment.}
\label{fig:case}

\end{figure}
\section{Patient Health Questionnaire-9 (PHQ-9)}

The Patient Health Questionnaire-9 (PHQ-9) is one of the most widely adopted 
clinical instruments for depression assessment and is frequently used as the 
primary supervision signal in multimodal depression recognition studies. 
Introduced as a self-administered screening tool aligned with DSM diagnostic 
criteria, PHQ-9 evaluates depressive symptoms through nine items, each scored 
from 0 to 3, producing a total score ranging from 0 to 27\footnote{
PHQ-9 severity levels are commonly categorized as minimal (0--4), mild (5--9),
moderate (10--14), moderately severe (15--19), and severe (20--27).
}. 
While PHQ-9 provides a validated measure of depression severity, its aggregated 
score alone does not reveal the behavioral evidence underlying the assessment. 
In clinical practice, mental health professionals interpret PHQ-9 scores together 
with observable psychological indicators, including affective expression, speech 
characteristics, cognitive responses, and social behaviors. 
This limitation motivates computational models that not only achieve accurate 
PHQ-9-based depression assessment but also provide interpretable behavioral 
evidence to support clinical decision-making.

\section{Implementation Details}
\label{app:more_implementation}

Following the standard Patient Health Questionnaire-9 (PHQ-9) protocol~\cite{kroenke2001phq}, subjects are classified as depressed or non-depressed using the clinical diagnostic threshold. Details of the preprocessing pipeline and label assignment are provided in the Appendix. Unless otherwise specified, identical training protocols and hyperparameter settings are adopted across all experiments.

\paragraph{PHQ-9-Based Clinical Threshold}
The Patient Health Questionnaire-9 (PHQ-9) score ranges from 0 to 27, where higher scores indicate more severe depressive symptoms. To transform the original PHQ-9 score into a normalized value between 0 and 1, a min-max normalization method is applied:
\begin{equation}
   X_{norm}=\frac{X_{PHQ9}}{27} 
\end{equation}

where $X_{PHQ9}$ represents the original PHQ-9 total score and ($X_{norm}$) represents the normalized depression severity score. After normalization, a score of 0 corresponds to the minimum symptom level, while a score of 1 corresponds to the maximum possible PHQ-9 severity.

For depression screening, a commonly used clinical cutoff is a PHQ-9 score of 10. Therefore, the normalized threshold can be calculated as:

\begin{equation}
    Threshold=\frac{10}{27}=0.370
\end{equation}

Accordingly, participants are classified into two groups based on the normalized score:

\begin{equation}
Y=
\begin{cases}
1, & X_{norm}\geq0.370 \\
0, & X_{norm}<0.370
\end{cases}
\end{equation}

where (Y=1) indicates a positive depression screening result (potential depressive symptoms), and (Y=0) indicates a negative screening result. We set the threshold at 0.370 as the clinical diagnostic threshold for depression soft labels.

For machine learning applications, the normalized PHQ-9 score can be used as a continuous severity indicator, while the binary label generated using the cutoff value can be used as the classification target. It should be noted that the normalized score represents symptom severity rather than a direct probability of having depression.

\section{Case Study}

To further examine the practical interpretability of BehavDep, Figure~\ref{fig:case} presents an example prediction report generated by the proposed framework. The report summarizes the model prediction together with the most influential behavioral concepts and their corresponding behavioral evidence. Two mental health professionals independently reviewed the extracted evidence and found that it was well aligned with the observed user behaviors.

In this case, BehavDep predicts a depressive state with a probability of 60.5\%, primarily supported by acoustic concepts. In particular, speech latency, long pauses, and reduced speech fluency are identified as salient evidence, which are summarized as behavioral patterns related to reduced speech energy and psychomotor slowing. Rather than providing only an overall prediction, BehavDep traces the prediction to specific and observable behavioral concepts, making the model output easier to interpret and inspect.

Overall, the case study shows how BehavDep can translate multimodal representations into human-readable behavioral evidence, providing a more transparent view of the patterns underlying individual predictions while leaving the final assessment to human professionals.

\end{document}